\documentclass{article}

\usepackage[nonatbib,preprint]{neurips_2025}

\usepackage[utf8]{inputenc} 
\usepackage[T1]{fontenc}    
\usepackage[colorlinks=true,allcolors=blue]{hyperref}       
\usepackage{url}            
\usepackage{booktabs}       
\usepackage{amsfonts}       
\usepackage{nicefrac}       
\usepackage{microtype}      
\usepackage{xcolor}         

\usepackage{graphicx} 
\usepackage{amsmath} 
\usepackage{algorithm}
\usepackage{algpseudocode}
\usepackage{float}

\newcommand{\R}{\mathbb{R}}
\newcommand{\vc}{\mathbf}

\newcommand{\eps}{\varepsilon}
\newcommand{\id}{\mathrm{d}}

\usepackage[numbers]{natbib}

\title{Sequential decoder training for improved latent space dynamics identification}

\author{%
  William Anderson \\
  Center for Applied Scientific Computing\\
  Lawrence Livermore National Laboratory\\
  Livermore, CA 94550 \\
  \texttt{anderson316@llnl.gov} 
  \And
  Seung Whan Chung\\
  Center for Applied Scientific Computing\\
  Lawrence Livermore National Laboratory\\
  Livermore, CA 94550 \\
  \texttt{chung28@llnl.gov}
  \And
  Youngsoo Choi \\
  Center for Applied Scientific Computing\\
  Lawrence Livermore National Laboratory\\
  Livermore, CA 94550 \\
  \texttt{choi15@llnl.gov} 
}

\begin{document}

\maketitle

\begin{abstract}
Accurate numerical solutions of partial differential equations are essential in many scientific fields but often require computationally expensive solvers, motivating reduced-order models (ROMs). 
Latent Space Dynamics Identification (LaSDI) is a data-driven ROM framework that combines autoencoders with equation discovery to learn interpretable latent dynamics.
However, enforcing latent dynamics during training can compromise reconstruction accuracy of the model for simulation data. 
We introduce multi-stage LaSDI (mLaSDI), a framework that improves reconstruction and prediction accuracy by sequentially learning additional decoders to correct residual errors from previous stages.
Applied to the 1D-1V Vlasov equation, mLaSDI consistently outperforms standard LaSDI, achieving lower prediction errors and reduced training time across a wide range of architectures.
\end{abstract}

\section{Introduction}

Simulating time-dependent partial differential equations (PDEs) is central to advances in engineering \cite{Calder2018,Cummings2015,Jones2020}, physics \cite{Thijssen2007,Vasileska2017}, and biology \cite{Noble2002}, but high-fidelity methods are often computationally expensive.
Reduced-order models (ROMs) address this issue by approximating the dynamics at far lower computational cost.
Here, we develop ROMs for PDEs with parametric dependence affecting the initial conditions or underlying physics.

Projection-based ROMs \cite{Benner2015, Berkooz1993,Bonneville2024a,Diaz2024,Kim2022,choi2021space,choi2020gradient,mcbane2021component,copeland2022reduced,choi2020sns,kim2021efficient,chung2024train} are generally interpretable, but require knowledge of the underlying governing equations or source code.
Non-intrusive ROMs are purely data-driven, but often lack interpretability.
To address this gap, \citet{Fries2022} introduced Latent Space Dynamics Identification (LaSDI), which combines an autoencoder with Sparse Identification of Nonlinear Dynamics (SINDy) \cite{Brunton2016} to learn interpretable dynamics in the compressed latent space. LaSDI generates ordinary differential equations (ODEs) for each training input parameter and interpolates these ODEs for unseen input parameters.

Despite many improvements and variations \cite{Bonneville2024,He2023,He2025,Park2024,Tran2024,he2025thermodynamically,chung2025latent,he2022certified,brown2023data}, LaSDI requires autoencoders to balance accurate data reconstruction with enforcing prescribed latent dynamics.
This compromise can lead to poor reconstruction and prediction accuracy.
Additionally, training remains costly due to large autoencoders and extensive hyperparameter tuning.
While multi-stage architectures exist in other contexts \cite{Aldirany2024,Wang2024,Howard2025,Vincent2010,Yu2022,Zabalza2016}, none of these methods address the interpretability-accuracy trade-off inherent to LaSDI and other models with an interpretable latent space \cite{Benner2020, Champion2019, Cranmer2020, Issan2022, Qian2020}.


To overcome these limitations, we introduce \textbf{multi-stage Latent Space Dynamics Identification (mLaSDI)}, a general framework that improves any LaSDI variant by sequentially reducing residual errors. 
Each additional stage learns a decoder which maps latent space trajectories to the residual error of previous stages, providing additional representation power while inheriting the interpretable latent space from the first stage. 
This allows us to significantly improve reconstruction and prediction accuracy, reducing the need for large autoencoders or increasing model robustness to hyperparameters.

\begin{figure}[t!]
  \centering
  \includegraphics[width = .8\linewidth]{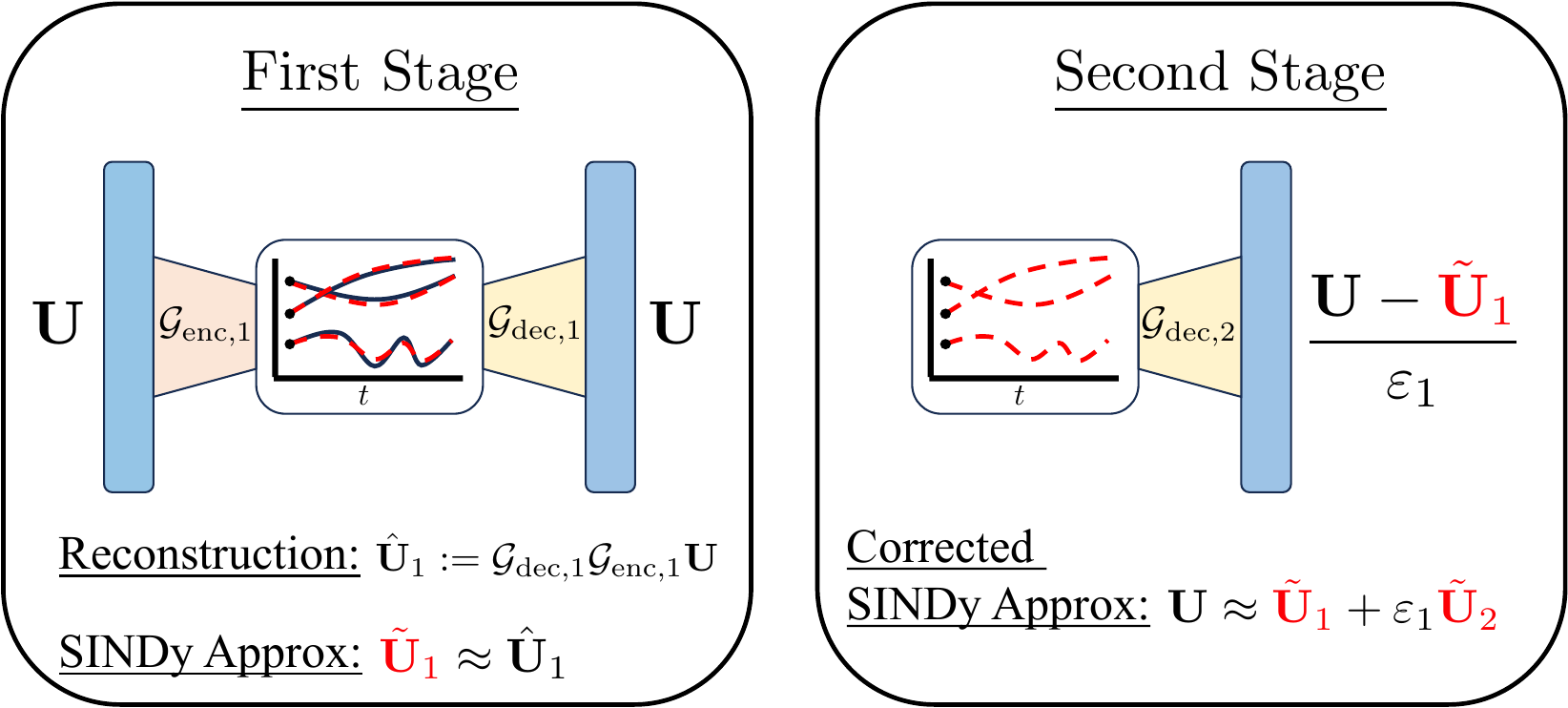}
  \caption{Schematic of mLaSDI. The first stage learns an autoencoder which reconstructs the training data $\mathbf{U}$, and latent space trajectories (solid black lines) are approximated using SINDy (dashed red lines). The second stage learns a new decoder which maps the SINDy trajectories in the latent space to the normalized residual error from the first stage.}
  \label{fig:mlasdi_scheme}
\end{figure}

\section{Multi-stage latent space dynamics identification}
\label{sec:Preliminaries}


Although mLaSDI is applicable to any variant of LaSDI, here we focus on comparisons to Gaussian Process-based LaSDI (GPLaSDI).
We briefly describe the GPLaSDI algorithm and refer the reader to \cite{Bonneville2024} for full details.

\paragraph{GPLaSDI} We consider dynamical systems of the form
\begin{equation}
	\frac{\id }{ \id t }\vc u (t; \pmb \mu) = \vc f (\vc u), \quad \vc u(0; \pmb \mu) = \vc u_0 (\pmb \mu),
	\label{eq:govode}
\end{equation}
where $\vc u : \R^{+} \to \R^{N_u}$ is the state vector, $\mathbf f: \R^{N_u} \to \R^{N_u}$ is a vector-valued function, and $\pmb \mu \in \mathcal D \subset \R^{N}$ is an input parameter affecting physics of the simulation or the initial condition. 
We assume no knowledge of the underlying dynamical system, and only consider snapshots of the state vector $\vc u$ taken at uniform times $t_i, \ i = 0, 1, ..., N_t$, for each of the $N_\mu$ training parameters.
Given an input parameter $\pmb \mu^{(i)}$, we form snapshots of the state vector into the training data matrix $U^{(i)} = [\vc u^{(i)}(0), \vc u^{(i)}(t_1), ... , \vc u^{(i)}(N_t) ] $. 
Concatenating the snapshots from each of our $N_{\mu}$ training parameters, we obtain the training data tensor $\mathbf{U} = [  U^{(1)},  U^{(2)}, ...,  U^{(N_{\mu})} ]\in \R^{N_{\mu} \times (N_t + 1) \times N_u}.$

After obtaining our snapshots, an autoencoder compresses the spatial dimension of the training data through encoder and decoder mappings $\mathcal{G}_{\text{dec}} : \R^{N_u} \to \R^{N_z}$ and  $ \mathcal{G}_{\text{enc}} : \R^{N_z} \to \R^{N_u}$, where $N_z \ll N_u$. 
The autoencoder reconstruction of the training data is defined by $ \hat{\mathbf U} = \mathcal{G}_{\text{dec}} \mathcal{G}_{\text{enc}}\mathbf U$, where the encoder and decoder minimize the training reconstruction loss $\mathcal L_{\text{AE}} ( \pmb \theta_\text{enc} , \pmb \theta_\text{dec} ) = \| \mathbf{U} - \hat{\mathbf{U}} \|^2$.  
Here  $\| \cdot \|$ is the element-wise $\ell^2$-norm and $\pmb \theta_\text{enc}, \pmb \theta_\text{dec}$ are model parameters of the encoder and decoder.

With the goal of obtaining an interpretable latent space, we consider the compressed data tensor $\mathbf{Z} = \mathcal{G}_{\text{enc}} \mathbf{U} \in \R^{N_{\mu} \times (N_t + 1) \times N_z}.$
For the snapshots of each compressed simulation $Z^{(i)}$, we follow \cite{Bonneville2024,He2023} and apply SINDy \cite{Brunton2016} to approximate dynamics of our latent space by
\begin{equation}
	\dot{Z}^{(i)} \approx \dot{\hat{Z}}^{(i)} := \Theta ( Z^{(i)} ) \Xi^{(i)}, \quad \Theta ( Z^{(i)} ) = (\vc 1 \ (Z^{(i)})^\top ), \quad \Xi^{(i)} = ( \vc b^{(i)} \  A^{(i)} )^\top,
\end{equation}
where we must learn the coefficients $\vc b^{(i)} \in \R^{N_z}$, and $A^{(i)} \in \R^{N_z\times N_z}$.
Defining the the SINDy coefficient tensor $\pmb \Xi = [  \Xi^{(1)} \  \Xi^{(2)} \ ... \   \Xi^{(N_{\mu})} ]\in \R^{N_{\mu} \times (N_z + 1) \times N_z}$, we determine our coefficients through minimizing the dynamics identification loss $\mathcal L_{\text{DI}} ( \pmb \Xi) = \| \dot{ \mathbf{Z}} - \dot{ \hat{ \mathbf{Z}} }\|^2$. 
Additionally, we penalize the norm of our SINDy coefficients to obtain the loss function for GPLaSDI
\begin{equation}
	\mathcal L ( \pmb \theta_\text{enc} , \pmb \theta_\text{dec}, \pmb \Xi ) = \mathcal L_{\text{AE}} ( \pmb \theta_\text{enc} , \pmb \theta_\text{dec} )  + \beta_1 \mathcal L_{\text{DI}} ( \pmb \Xi) + \beta_2 \| \pmb \Xi \|^2,
    \label{eq:lasdiloss}
\end{equation}
where $\beta_1, \beta_2$ are hyperparameters.

After training our model, we use the GP interpolation scheme of GPLaSDI \cite{Bonneville2024} to interpolate SINDy coefficients $\Xi^{(*)}$ for a new test parameter $\pmb \mu^{(*)}$. 
Essentially, we perform Gaussian Process Regression on each of the SINDy coefficients in the training set to find $\mathcal{GP}_{\pmb \Xi}: \pmb \mu^{(*)} \mapsto \{m( \Xi^{(*)}), ( \Xi^{(*)})\}$, where $m$ is the predictive mean of $ \Xi^{(*)}$ and $s$ the predictive standard deviation.
This allows us to obtain a mean prediction, but we can also sample the GP to get uncertainty for our predictions.
%
\begin{table}[t]
  \caption{Model architectures used to train GPLaSDI and mLaSDI for 1D-1V Vlasov experiments.}
  \label{tab:1d1v}
  \centering
  \footnotesize
  \begin{tabular}{lll}
    \toprule
    \textbf{Component} & \textbf{Choices} & \textbf{Description} \\
    \midrule
    mLaSDI Hidden Layers & 50, 500, 1000, 500-50, 1000-500-50 & Fully connected layers \\ 
    GPLaSDI Hidden Layers & 75, 750, 1500, 750-75, 1500-750-75 & Fully connected layers \\ 
    Latent Dimension & 4, 5, 6, 7 & Bottleneck dimension \\ 
    GPLaSDI Training Config & 25k, 50k, 75k, 100k & Training checkpoints \\
    mLaSDI Training Config. & (25k,25k), (25k,50k), (50k,50k), (75k,25k) & (Stage 1, Stage 2) iter. pairs \\
    \bottomrule
  \end{tabular}
\end{table}
\paragraph{mLaSDI}  The key limitation of LaSDI is that training with loss function~\eqref{eq:lasdiloss} limits the representation power of our autoencoders.
In some cases, autoencoders trained by LaSDI fail to accurately reconstruct even the training data due to constraints imposed by our latent dynamics.
This is the main motivation for mLaSDI, where we propose training additional decoders to improve reconstruction accuracy when one stage of LaSDI fails to provide acceptable accuracy.
Decoders introduced after the first stage are not required to learn additional SINDy coefficients, but provide additional representation power while still inheriting the interpretable latent space from the first stage of LaSDI.

For training data $\mathbf{U}$, we obtain the autoencoder reconstruction of the data from the first stage $\hat{ \mathbf{U}}_1$.
We can also obtain SINDy reconstructions of the training data by decoding our SINDy approximations of the latent space dynamics, i.e. $\tilde{\mathbf{U}}_1 := \mathcal{G}_{\text{dec},1}({\hat{\mathbf{Z}}} )$, where the subscript refers to the fact that this is our first stage of training and ${\hat{\mathbf{Z}}}$ is our SINDy approximation of the latent trajectories ${{\mathbf{Z}}}$.

To improve our SINDy reconstruction accuracy, we consider the residual between training data ${\mathbf{U}}$ and the first stage SINDy reconstructions of the training data $\tilde{\mathbf{U}}_1$, given by $\vc R_1 = {\mathbf{U}} - \tilde{\mathbf{U}}_1$.
With mLaSDI, we introduce a second decoder which takes the first stage SINDy latent space trajectories as input, and attempts to reconstruct the (normalized) residual error. 
More precisely, we fix model parameters of the first autoencoder, and learn a mapping for the second stage $ \mathcal{G}_{\text{dec},2} : \R^{N_z} \to \R^{N_u}$ by minimizing
\begin{equation}
    \mathcal L_{\text{dec},2} ( \pmb \theta_{\text{dec},2} ) =  \| \eps_{1}^{-1} \vc R_{1} - \mathcal G_{\text{dec},2} \hat{\mathbf{Z}} \|^2, \quad \eps_1 = \mathrm{std}(\mathbf R_1).
\end{equation}
With this approach we can sequentially train multiple models to approximate the training data $\vc U$ as
\begin{equation}
    \vc U \approx \tilde{\mathbf{U}}_{1} + \eps_{1} \tilde{\mathbf{U}}_{2} + \eps_{2}\tilde{\mathbf{U}}_{3} + ... + \eps_{n-1}\tilde{\mathbf{U}}_{n},
\end{equation}
where $\tilde{\mathbf{U}}_{k}$ is the residual approximation at the $k^\mathrm{th}$ stage $\tilde{\mathbf{U}}_{k} = \mathcal G_{\text{dec},k} \hat{\mathbf{Z}}$, and $\eps_{k}$ normalizes the $k^\mathrm{th}$ residual to have standard deviation 1.
A schematic of mLaSDI describing this process is provided in Figure \ref{fig:mlasdi_scheme}.

We are free to choose any architecture for the decoders learned after the first stage.
Here, we take an approach similar to \citet{Wang2024} and use identical decoder architectures at all stages, varying only the activation functions.
After the first stage, we use sine activation on the first layer which helps to capture high-frequency components of the residual.
Any subsequent layers use hyperbolic tangent activation function.

\section{Results and discussion}
\label{sec:example}

\begin{figure}
  \centering
    \includegraphics[width= 0.9\linewidth]{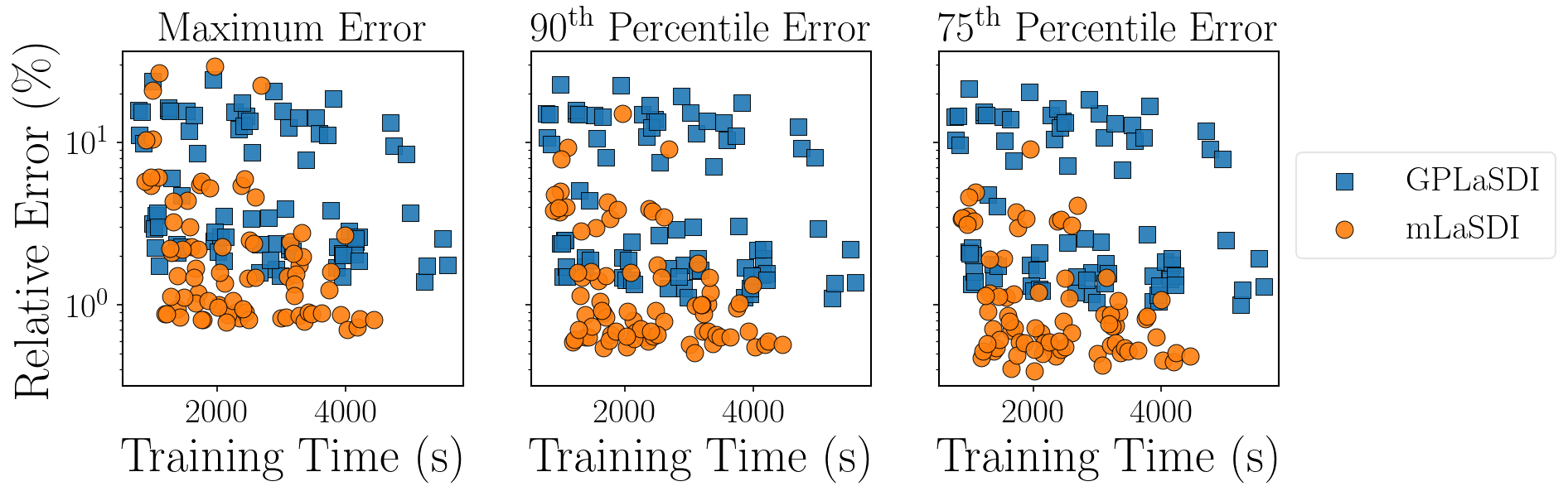}
  \caption{Applying GPlaSDI and mLaSDI with two stages to 1D-1V Vlasov equation using a wide range of model architectures in Table~\ref{tab:1d1v}. Percentile errors for the first and second stages using errors from~\eqref{eq:maxerrs}.}
  \label{fig:1d1v_std}
\end{figure}

We compare mLaSDI and GPLaSDI on the 1D-1V Vlasov equation
\begin{equation}
	\begin{cases}
		u_t +  (vu)_x +  (\Phi_x u)_v = 0, \quad t \in (0, 5], \ x \in [0, 2\pi], \ v \in [-7, 7], \\ 
		\Phi_{xx} = \int_v u \ \id v, \\
		u(x, v, 0; \pmb \mu) =  \frac{8}{\sqrt{2 \pi T}} \left[ 1 + 0.1\cos(kx) \right] \left[ \exp \left( - \frac{(v - 2)^2}{2T} \right) + \exp \left( - \frac{(v + 2)^2}{2T} \right) \right],
	\end{cases}
	\label{eq:Vlasov}
\end{equation}
where $u(x, v, t)$ is the plasma distribution function which depends on space $x$ and velocity $v$. 
The function $\Phi(x)$ is the electrostatic potential. 
The input parameter vector $\pmb \mu = [T, k]^\top$ controls the spread $T$ and periodicity $k$ of the initial two-stream configuration.

\paragraph{Data Generation} We solve \eqref{eq:Vlasov} using HyPar \cite{Hypar} with a WENO spatial discretization scheme \cite{Jiang1996} and RK4 time integration scheme with timestep $\Delta t = 0.005$. 
We run full-order simulations for parameter values $T \in [0.9, 1.1]$ and $k \in [1.0, 1.2]$, where the parameter ranges are discretized by $\Delta T = \Delta k = 0.01$.
We sample the solution at every timestep on a $64 \times 64$ grid in the space-velocity field to obtain 251 snapshots with state dimension $N_u =  4096$.

\paragraph{Training Details} We train GPLaSDI and two stages of mLaSDI, varying the number of training iterations, hidden layers, and bottleneck dimension for each experiment, as summarized in Table~\ref{tab:1d1v}.
The hidden layers for GPLaSDI were widened by a factor $1.5$ compared to mLaSDI, demonstrating that the multi-stage training achieves superior performance with fewer model parameters.
For each experiment we uniformly sample 25 points from the 2D parameter space and predict on the full set of 441 simulations.
All models are implemented in Pytorch and trained on a NVIDIA V100 (Volta) GPU using the Adam optimizer \cite{Kingma2014} with a learning rate of $10^{-3}$.
The mLaSDI loss weights are set to $\beta_1 = 0.1$, $\beta_2 = 0.001$.
SINDy coefficients are interpolated via scikit-learn’s \texttt{GaussianProcessRegressor} with a Mat\'{e}rn kernel \cite{sklearn}, and we only consider the mean predictions of the GPs.

\paragraph{Results} Our error metric is the maximum relative error between a full-order simulation $\vc u(t ; \pmb \mu^{(*)})$ and the mLaSDI approximation $\tilde{ \vc u}(t ; \pmb \mu^{(*)} )$, defined as 
\begin{equation}
    r^{(*)} := \max_{\substack{j = 0,...,N_t}} \frac{ \| \vc u(t_j; \pmb \mu^{(*)}) - \tilde{ \vc u}(t_j;\pmb \mu^{(*)}) \| }{ \| \vc u(t_j ; \pmb \mu^{(*)}) \| }.
    \label{eq:maxerrs}
\end{equation}
We then calculate the maximum relative errors for every parameter value $\{r^{(i)}\}_{i = 1}^{N_\mu}$ in our test set.
In Figure \ref{fig:1d1v_std} we plot the maximum, $90^\textrm{th}$, and $75^\textrm{th}$ percentile errors from each autoencoder, along with training time.
The reported training times for the mLaSDI represent the total time to train both the first and second stages.

We see that mLaSDI consistently produces lower relative errors than GPLaSDI models that have more model parameters.
Introduction of a second stage also allows us to achieve maximum relative errors below 1\% for some architectures, which we never achieve by training only one stage of GPLaSDI.
Across all tested models, mLaSDI provides a median increase in accuracy by a factor of 2.54, 2.78, and 3.06 for the maximum, $90^\textrm{th}$, and $75^\textrm{th}$ percentile errors, respectively.
These results demonstrate that mLaSDI maintains high accuracy across a wide range of model choices, reducing sensitivity to hyperparameters while improving prediction accuracy.
Consistent improvement in accuracy across different architectures demonstrates that mLaSDI's multi-stage approach addresses a fundamental limitation in LaSDI methods more effectively than architectural modifications alone. 
Our results suggest that the interpretability-accuracy trade-off is better resolved through training methodology rather than model scaling. 
This framework opens promising directions for enhancing other interpretable latent space methods and warrants investigation across broader PDE families.

\begin{ack}
This work was partially supported by the Lawrence Livermore National Laboratory (LLNL) under Project No. 50284. Y.\ Choi was also supported for this work by the U.S. Department of Energy, Office of Science, Office of Advanced Scientific Computing Research, as part of the CHaRMNET Mathematical Multifaceted Integrated Capability Center (MMICC) program, under Award Number DE-SC0023164.
Livermore National Laboratory is operated by Lawrence Livermore National Security, LLC, for the U.S. Department of Energy, National Nuclear Security Administration under Contract DE-AC52-07NA27344. 
LLNL document release number: LLNL-CONF-2010449.
\end{ack}

\bibliography{references.bib}

\end{document}